\documentclass[letterpaper, 10 pt, conference]{ieeeconf}  % Comment this line out if you need a4paper

\IEEEoverridecommandlockouts                              % This command is only needed if 
\usepackage{graphicx}
\usepackage{xcolor}
\usepackage[export]{adjustbox}
\usepackage{adjustbox}
\usepackage{lscape}
\usepackage{subfigure}
\usepackage{dingbat}
\usepackage{multirow}
\usepackage{multicol}
\usepackage{makecell}
\usepackage{hyperref}
\usepackage{url}
\newcommand\T{\rule{0pt}{2.9ex}}       % Top strut
\newcommand\B{\rule[-1.2ex]{0pt}{0pt}} % Bottom strut

\usepackage{flushend}  %just to make references appear nicely in two col.

\title{\LARGE \bf
Deep Learning-Enhanced Visual Monitoring in Hazardous Underwater Environments with a Swarm of Micro-Robots
}

\author{Shuang Chen, Yifeng He, Barry Lennox, Farshad Arvin and Amir Atapour-Abarghouei% <-this % stops a space
\thanks{This work was partially supported by the UK EPSRC RAIN, Robotics \& Artificial Intelligence for Nuclear [grant numbers EP/R026084/1 and EP/W001128/1].}% <-this % stops a space
\thanks{S. Chen, F. Arvin and A. Atapour-Abarghouei are with the Department of Computer Science, Durham University, Durham, UK
        {\tt\small shuang.chen@durham.ac.uk}}%
\thanks{Y. He and B. Lennox are with the Department of Electrical \& Electronic Engineering, The University of Manchester, Manchester, UK.%
}% <-this % stops a space
\thanks{Code and dataset can be found at: {\url{https://github.com/ChrisChen1023/Micro-Robot-Swarm}}
}
}

\begin{document}

\maketitle
\thispagestyle{empty}
\pagestyle{empty}

%%%%%%%%%%%%%%%%%%%%%%%%%%%%%%%%%%%%%%%%%%%%%%%%%%%%%%%%%%%%%%%%%%%%%%%%%%%%%%%%
\begin{abstract}
Long-term monitoring and exploration of extreme environments, such as underwater storage facilities, is costly, labor-intensive, and hazardous. Automating this process with low-cost, collaborative robots can greatly improve efficiency. 
These robots capture images from different positions, which must be processed simultaneously to create a spatio-temporal model of the facility. In this paper, we propose a novel approach that integrates data simulation, a multi-modal deep learning network for coordinate prediction, and image reassembly to address the challenges posed by environmental disturbances causing drift and rotation in the robots' positions and orientations. Our approach enhances the precision of alignment in noisy environments by integrating visual information from snapshots, global positional context from masks, and noisy coordinates. We validate our method through extensive experiments using synthetic data that simulate real-world robotic operations in underwater settings. The results demonstrate very high coordinate prediction accuracy and plausible image assembly, indicating the real-world applicability of our approach. The assembled images provide clear and coherent views of the underwater environment for effective monitoring and inspection, showcasing the potential for broader use in extreme settings, further contributing to improved safety, efficiency, and cost reduction in hazardous field monitoring.
% \textcolor{red}{Chris - I have added a bit - check, change and add please}. 

\end{abstract}

%%%%%%%%%%%%%%%%%%%%%%%%%%%%%%%%%%%%%%%%%%%%%%%%%%%%%%%%%%%%%%%%%%%%%%%%%%%%%%%%
\section{Introduction}

% general
Monitoring and measuring environmental conditions are essential in extreme environments, such as those characterised by high temperatures, radiation or underwater settings. Deploying human operators in such challenging conditions is often impractical or poses significant risks. Consequently, robotic systems present a safer and more reliable alternative for conducting these critical missions.
For instance, in nuclear power plants, nuclear waste is stored in pools of water known as spent fuel ponds contained within specially designed rods. To ensure the safety and integrity of these storage sites, specialised underwater camera systems monitor the condition, position, and quantity of the nuclear waste.
The International Atomic Energy Agency (IAEA) allocates substantial resources—over £25 million annually~\cite{pepper2001lessons} to inspect nuclear fuel waste storage using the manual IAEA DCM-14 camera~\cite{doyle2011nuclear}. However, regularly inspecting the ponds is time-consuming and repetitive, highlighting the need for more efficient monitoring solutions, such as automated robotic systems.

%- single robot 
While robots with specialised sensory systems can automate inspection, relying on a single-robot configuration, in large and uncharted environments, can be a significant concern. Consequently, deploying multiple robots can enhance coverage, robustness, and effectiveness in these settings~\cite{dorigo2020reflections}.

\begin{figure}[tb]
\centering
\includegraphics[width=0.45\textwidth]{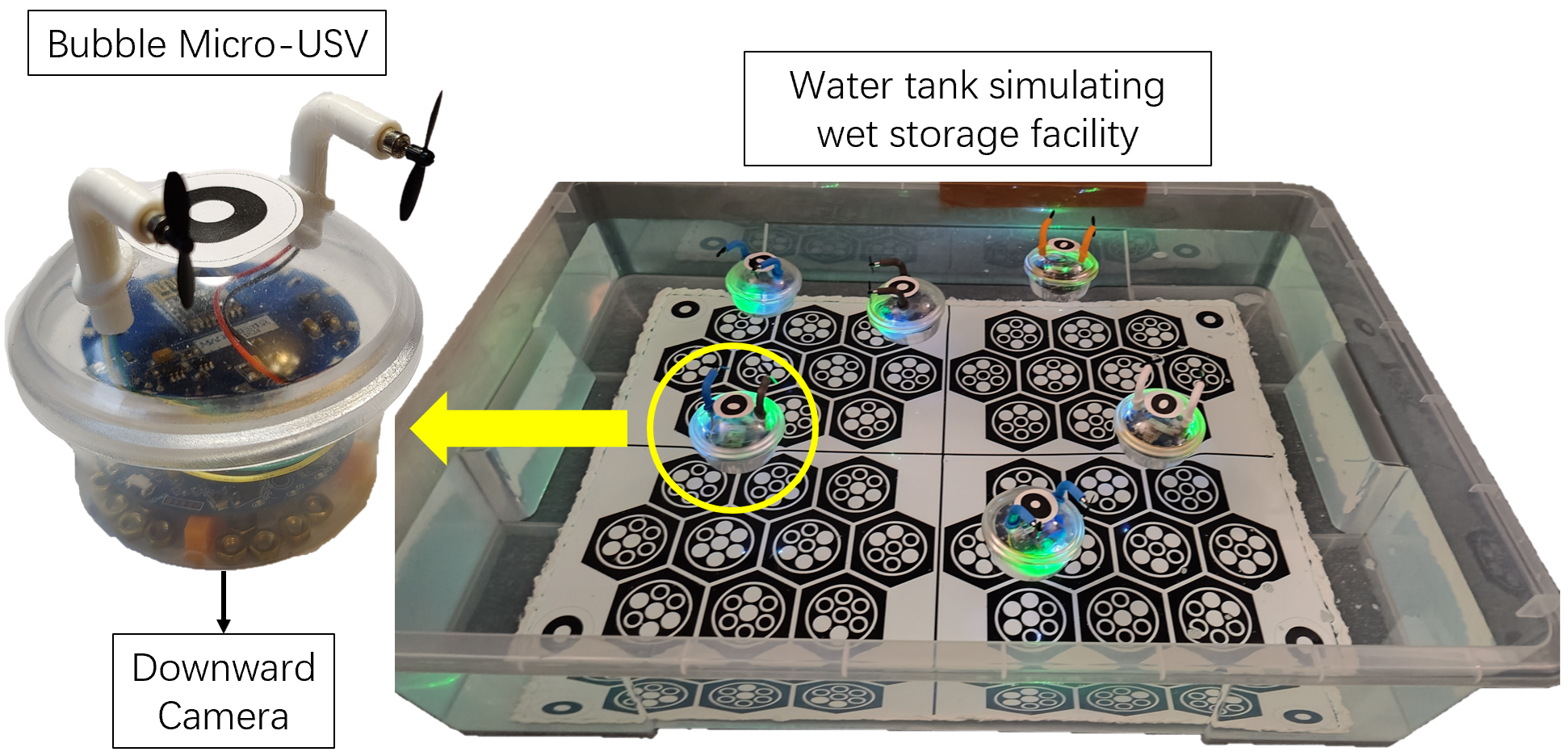}
\caption{The overarching vision of the proposed visual exploration system involves deploying a swarm of micro-surface robots,  \textit{Bubbles}.%, to conduct visual inspections of underwater storage facilities. %These robots collaboratively capture images from various positions across the pond. %By integrating and stitching together 100 individual images collected by the swarm, the system aims to generate a comprehensive panoramic representation of the facility.
}
\label{concept-fig1}\vspace{-0.4cm}
\end{figure}

% \textcolor{red}{\textbf{Chris-Amir-Farshad  will Write Here: }}
To enable a full inspection of the scene monitored by underwater robots, the snapshots captured the robots must be assembled into a cohesive view. However, this task is complicated by the nature of the environment. The movement of the liquid in the pond causes the robots to drift unpredictably, which introduces random and irregular noise into positional coordinates and rotation angle information recorded during snapshot capture. As a result, attempts to directly stitch these noisy snapshots produce confusing and misaligned images, which do not provide a clear understanding of the scene (as the noisy stitched image shown in Fig.~\ref{fig:comparison}).

In this paper, we propose a pipeline that includes data simulation, a multi-modal deep learning network for coordinate prediction, and image reassembly. A novel approach is introduced for generating synthetic images, that simulate spent fuel ponds, described in Sec.~\ref{sec:img_synthetic}. Additionally, we present a dataset SFP10 with 11,000 sets of disturbed data, simulating the drifting effects experienced by robots navigating the fuel pond. The code and dataset are publicly available.
% \url{https://github.com/ChrisChen1023/Micro-Robot-Swarm}

The contributions in our work are summarised as:
\begin{itemize}
    \item \textbf{Robotics:} We propose a robust, low-cost methodology for coordinating robotic swarm in underwater environments, ensuring effective image capture and assembly despite disturbances such as drifting and rotation.
    \item \textbf{Pipeline:} We introduce an integrated pipeline that combines data simulation, coordinate prediction through a multi-modal deep learning network, and robust image assembly, significantly improving the precision of alignment in noisy environments.
    \item \textbf{Dataset:} We create and release a comprehensive dataset SFP10 with 11,000 sets of disturbed data to simulate real-world robotic operations in challenging underwater scenes, providing a valuable resource for future research.
\end{itemize}

% We propose a novel multi-modal framework to predict accurate coordinates for underwater robotic systems, specifically addressing the challenges posed by noise in real-world robotic operations. The framework integrates visual information from snapshots, global positional context from masks, and noisy coordinates to enhance the model's ability to predict accurate coordinates. This approach allows for robust prediction of both the position and orientation $(x,t,\theta)$ of the robots, enabling the creation of well-aligned full images from multiple snapshots.

\section{Related Work}

We consider related work in Robotic Visual Monitoring (Sec. \ref{sec:related:robots}) and Image Assembly (Sec. \ref{sec:related:assembly}). 

\subsection{Visual Monitoring by Micro Robots}
\label{sec:related:robots}

While prior work~\cite{lennox2019embodiment,huang2019exploration,west2019debris} has focused on robotic inspection of storage facilities, the implementation of such automated systems presents significant challenges. For instance, ~\cite{groves2019mallard} developed \textit{MallARD}, an autonomous aquatic surface vehicle specifically designed for monitoring nuclear storage ponds. MallARD is equipped with four thrusters that enable navigation across the two-dimensional water surface and includes a camera in its payload area for conducting underwater inspections. Similarly, the \textit{MASKI+} robotic system \cite{provencher2016maski+} uses a controller that provides five degrees of freedom in water for diagnostic and intervention tasks in hydroelectric power plants. We also developed \textit{AVEXIS}~\cite{griffiths2016avexis}, a low-cost micro-submersible designed to monitor nuclear underwater storage facilities with limited access points. In a subsequent deployment~\cite{nancekievill2018development}, we equipped the robot with radiological sensors for inspecting the Fukushima Daiichi site in Japan.

For multi-robot examples, ~\cite{green2021minimalist} introduced a multi-robot coverage problem in a barrier-laden environment using the Pioneer P3-DX robot platform and a simulated barrier coverage robot. Similarly, in~\cite{patel2024multi}, multiple robots  were developed for inspecting oil and gas pipelines through a wireless autonomous surface vehicle (ASV) relay, enabling coordinated operations.~\cite{xanthidis2022multi} employed an ASV named \textit{Aqua2} for underwater exploration, specifically for mapping and monitoring shipwrecks. Despite the results of prior work, generating an accurate long-term spatio-temporal model of these facilities remains the main challenge that needs to be addressed. Our approach can assemble snapshots captured by micro robot swarms to allow for a complete live inspection of the facility.

\subsection{Image Reassembly}
\label{sec:related:assembly}

Recent advances in image reassembly (reassembling images from disjointed fragments) have explored unsupervised and self-supervised learning approaches. For instance, \cite{zhang2014graph} integrates geometry and colour information for image reassembly by pairwise matching of fragment boundaries, global image reassembly through a graph-based search algorithm and refinement of the reassembly using a graph optimisation technique \cite {kummerle2011g} to reduce accumulated errors. Learning-based approaches have also been used for image reassembly. For example, \cite{li2021jigsawgan} integrates both boundary and semantic information to improve puzzle reconstruction. This multi-task pipeline incorporates a branch for predicting jigsaw permutations and another branch for generating images \cite{goodfellow2014generative} with the correct order. \cite{song2023siamese} combines reinforcement learning \cite{mnih2015human} with Siamese networks \cite{chicco2021siamese} to optimise fragment swapping to correctly reassemble puzzles.

Image assembly and puzzle solving methods have also played a significant role in enhancing the performance of other tasks. \cite{noroozi2016unsupervised} introduces a self-supervised approach for learning image representations by solving jigsaw puzzles as a pretext task, to allow the network to learn object parts and their spatial arrangement. Puzzle-solving \cite{salehi2020puzzle} has also been used to enhance anomaly detection in images. JiGen \cite{carlucci2019domain} leverages jigsaw puzzle-solving for domain generalisation by jointly learning spatial correlations and object classification. \cite{wei2019iterative} uses weak spatial cues to iteratively solve jigsaw puzzles by combining unary and binary terms to assess the likelihood of patches being correctly positioned relative to one another.

Despite the advances in image reassembly techniques, none of these methods are equipped to handle the unique challenges of rotation and overlap that appear in our application. In the context of our work, fluid dynamics may cause the robots to move and rotate unpredictably. The precise amount of rotation before each snapshot is captured can vary, as the robots are subject to water currents, buoyancy, and other forces. This makes it challenging to reconstruct images when the degree of rotation is uncertain and not accounted for. Additionally, many of the snapshot overlap, complicating the alignment process. Many existing methods \cite{zhang2014graph, li2021jigsawgan} assume fixed orientations and non-overlapping images and are, as such, unsuitable in our application. Consequently, we have developed a novel pipeline to address these specific challenges to ensure robust image reassembly even with rotational variation and overlapping image regions.

\section{Micro-Robotic Platform}
\begin{figure}[b]
\centering
\includegraphics[width=0.48\textwidth]{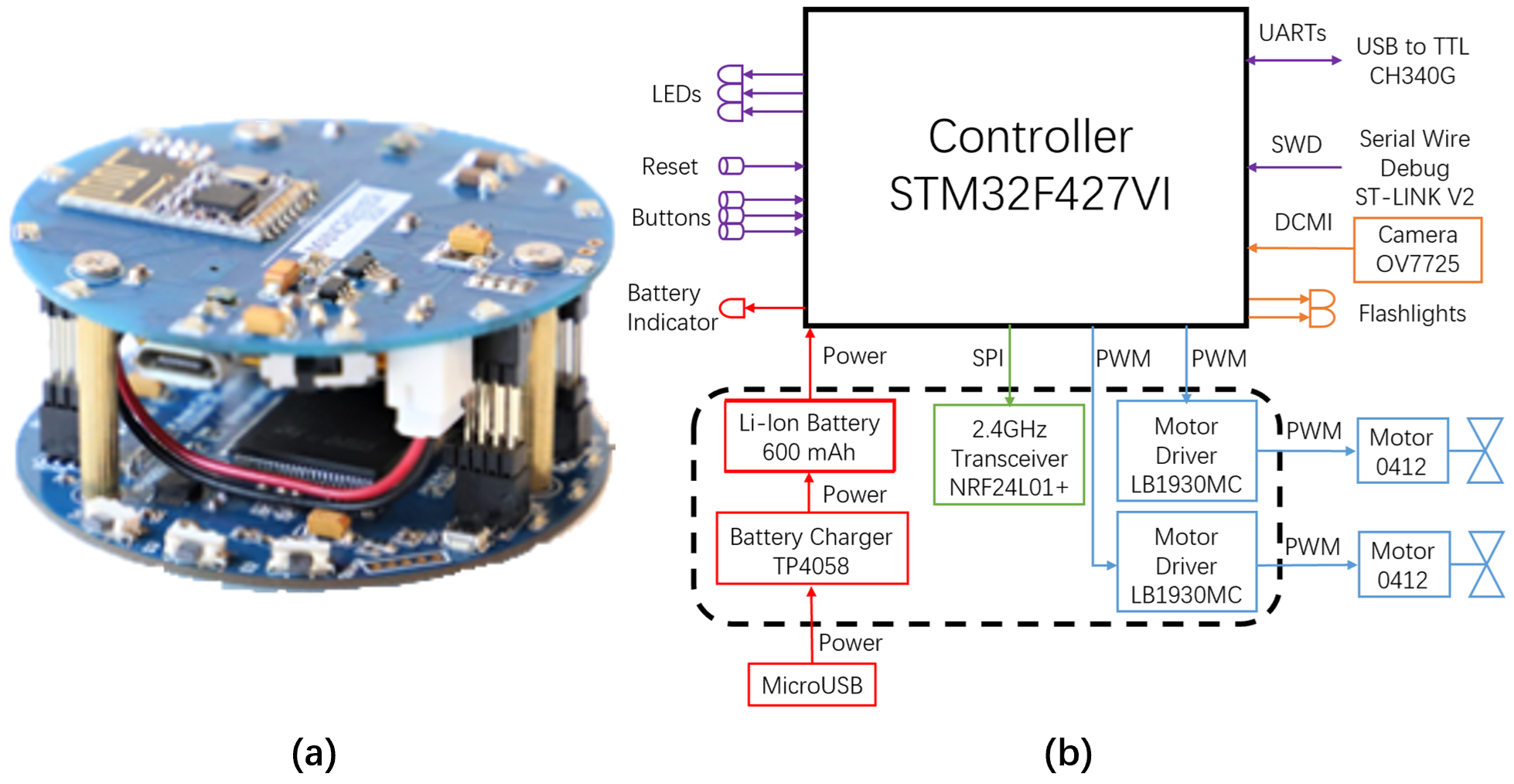}
% \vspace{-0.1cm}
\caption{(a) The electronics of Bubble, divided into two PCBs with different functionalities. (b) The system architecture of the robot illustrates its key functions. The section within the dashed rectangle represents the top PCB.}
\label{fig-robot-arch}
\end{figure}

In this section, we briefly introduce main components of the \textit{Bubble} micro surface robot which has been developed for perpetual monitoring of wet storage facilities~\cite{he2024bubbles}. Figure~\ref{fig-robot-arch} illustrates robot's hardware module and architecture. 
%\vspace{-0.25cm}
%%%% photo of Bubble 

\subsection{Bubble Robot Specification}
An STM32F427VI microcontroller manages low-level tasks, such as image capture, actuation, and communication. The  low-power microcontroller is a Cortex-M4 single-core 32-bit processor operating at 180~MHz with 0.25~MB SRAM and 2~MB Flash memory embedded with a Floating Processing Unit~(FPU), which enables the real-time image processing. An NRF24L01+ wireless communication module is attached to the main board of Bubble to transfer images to the base station. The robot uses a compact low-cost OV7725 camera with a maximum resolution of 640$\times$480 pixels at 30~fps. 160$\times$110-pixel images are used here to optimise memory usage and communication bandwidth. The captured image goes into an external buffer, but the robot can store eight images locally. This can be extended by reducing the image size and increasing the memory size.

As shown in Fig.~\ref{concept-fig1}, the Bubble enclosure is 3D printed out of plastic. Two actuators on top reduce cross contamination from storage facilities. The enclosure is a waterproof transparent case allowing camera to see downward and capture images from the bottom of the pond. The Bubble is equipped with two coreless DC motors as its actuators. %It is capable of achieving a high rotation speed of up to 53000~RPM. 
The motors are paired with 2.1~cm diameter twin-wing plastic propellers commonly used in tiny drones. %Two LB1930MC motor driver chips on the top PCB control the propulsion system designed to drive the motor with enhanced power.

\subsection{Experimental Setup}

The experimental simulation pond is 75$\times$52~cm. Each robotic unit is equipped with circular markers, with a diameter of 3~cm each. The hexagonal patterns and the solid and hollow circles on the poster at the bottom of the simulation pond (Fig.~\ref{concept-fig1}) are designed to replicate the layout and features of an actual spent fuel pond in nuclear power stations.

% aes model
An efficient swarm coordination mechanism is essential for conducting such exploration tasks. Collective Motion, a standard swarm behaviour, offers a promising approach for real-world applications by enabling coordinated movement and operation of multiple agents. In this work, we used the state-of-the-art collective motion mechanism that is based on the active elastic sheet framework~\cite{ferrante2013collective}. It places virtual springs between adjacent robots, where both distance and angular differences between robots generate repulsive and attractive forces to enable the robots to remain aligned and achieve coordinated collective motion.

\begin{figure}[tb]
\centering
\includegraphics[width=0.45\textwidth]{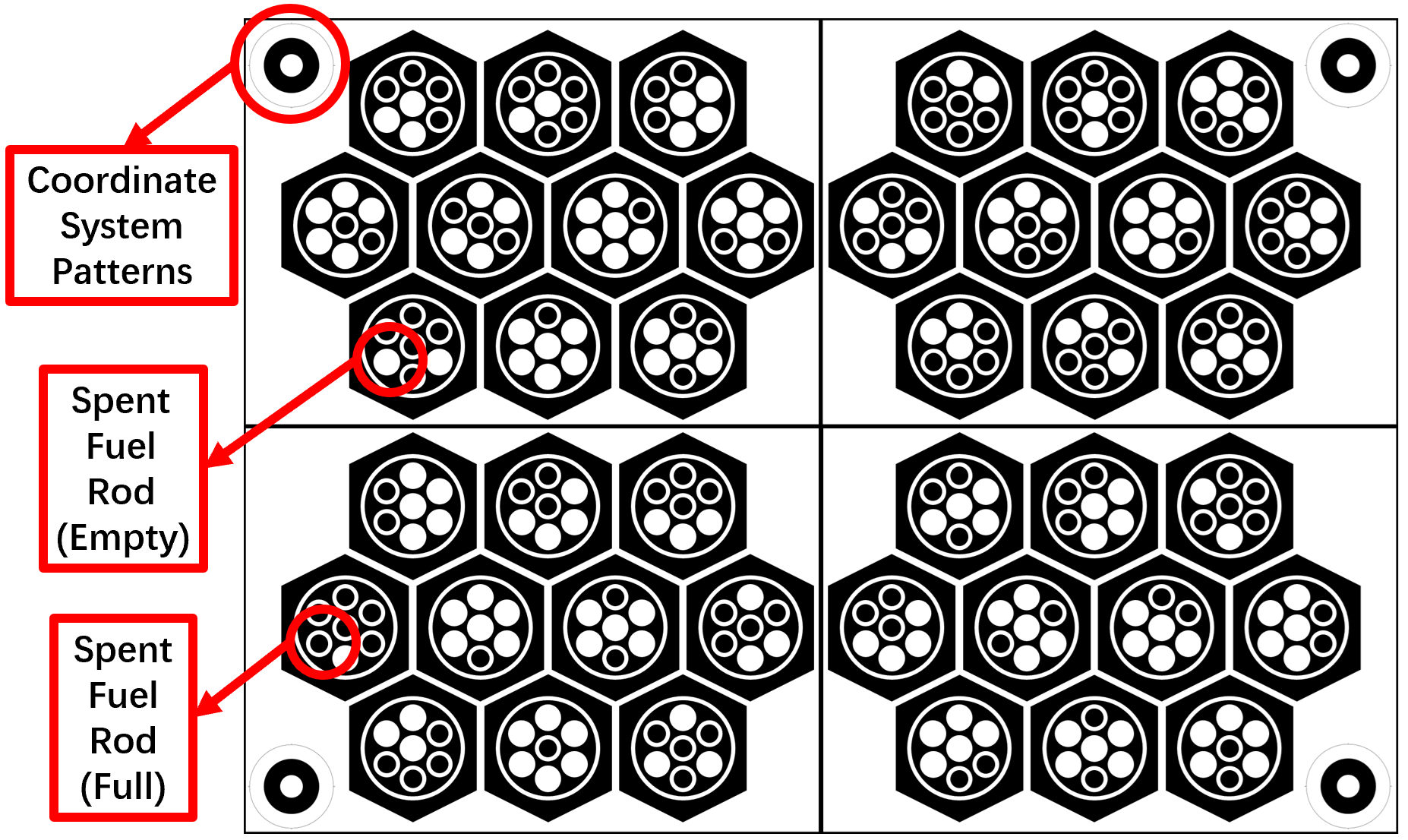}
\caption{Image simulating the spent fuel pond; hollow circles indicate empty rods and solid circles indicate full rods.}
\label{figE1}\vspace{-0.4cm}
\end{figure}

Whycon \cite{whycon_icar} is used with a low-cost USB camera to localise and track the robots. The camera captures the entire area from an overhead perspective. Each Bubble robot is equipped with a circular marker to enable tracking. %Additionally, circular markers are placed at each corner of the target poster to establish a coordinate system for the surface of the simulation pond. The Bubbles operated within this defined framework, measuring 54$\times$37~cm. 
Due to hardware limitations, only a single Bubble robot can communicate simultaneously with the other six Bubbles. In this configuration, one robot acts as the leader, while the other six serve as followers.

% \vspace*{0.4cm}
\begin{figure*}[t]
\vspace*{0.4cm}
% dUJJJJJJJJJJJJJJJJJJUJYU  VV  
\centering
  \includegraphics[height=6cm]{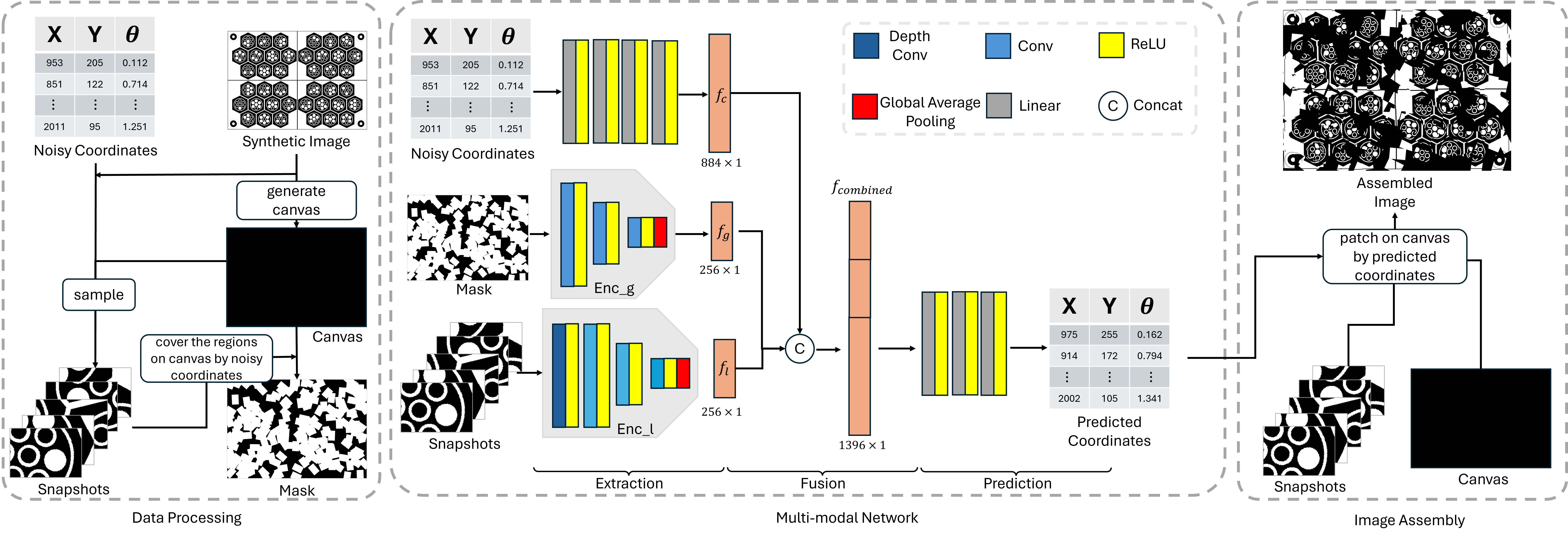}
  \vspace{-0.2cm}
  \caption{Overview of the overall pipeline, which integrates data simulation (Left), noisy coordinate correction via a deep learning network (Middle), and image reassembly (Right) to generate coherent images from snapshots captured by micro-robots in noisy underwater environments.}
  \vspace{-0.6cm}
  \label{fig:overview}
\end{figure*}

\section{Dataset}
\label{sec:data}
Since the learning-based model in charge of predicting correct robot coordinates requires a significant amount of data to be trained, capturing real-world data using the simulation pond is not feasible. As such, we require synthetic pond images, snapshot masks and noisy coordinates of the robot to train and evaluate our model. The model trained on the synthetic data will predict accurate coordinates by combining visual information from snapshots, spatial context from masks, and positional data from noisy coordinates.

In this work, we present a synthetic dataset, SFP10, designed to simulate data captured from spent fuel ponds. The dataset consists of 10,000 images for training and 1,000 for testing. Each set includes synthetic pond images, snapshots, perturbed positional coordinates as well as their corresponding binary masks. The synthetic pond images replicate the layout of a spent fuel pond, while snapshots capture various robot viewpoints and orientations as they navigate the environment. The perturbed positional coordinates are added to simulate real-world drift and rotational noise, which are inherent in such environments. The binary masks provide spatial context by explicitly marking the regions covered by the snapshots, facilitating accurate prediction of the robots' positions and orientations. These components are detailed in the following.

\subsection{Pond Image Synthesis}
\label{sec:img_synthetic}
Each generated image represents a top-down view of fuel rods submerged in water, with variations in the internal structure of the rods (as shown in Fig.~\ref{figE1}). Specifically, the inner circles of each rod are either white or partially filled. We assign a 0.5 probability for each inner circle to be fully or partially filled, while other characteristics, such as size, arrangement, and position of the rods, remaining fixed. Using this method, we generate 100 base images for use.

\subsection{Snapshot Acquisition}
\label{sec:snapshot_generation}
After generating the base pond images, we simulate the capture process carried out by the swarm of micro-surface robots. The Bubbles are designed to move across the pond and capture images from various positions ($x$,$y$) and orientations ($\theta$), representing their rotational angle. From each of the 100 base images, we randomly sample 221 snapshots with unique ($x$,$y$,$\theta$) values, simulating different viewpoints and rotations. Each snapshot is set to a fixed size of 160$\times$110 pixels, matching the resolution in the physical setup. This process produces a diverse set of snapshots, mimicking the behaviour of the robots as they move around the pond.

\subsection{Perturbed Coordinate Data}
\label{sec:add_noise}
To simulate real-world imperfections during image capture, we introduce noise into the data. Noise is sampled from a normal distribution and added to $x$, $y$ and $\theta$, representing the effects of drifting and random rotation that could occur as the robots navigate the pond. This step simulated the small perturbations that might occur due to environmental factors or mechanical drift in the robots’ movement. For each base image, we apply 100 different noise instances, for a total dataset of 10,000 sets of coordinates.

\subsection{Region Mapping with Mask}
\label{sec:mask_generation}
\begin{figure*}[t] \centering
\vspace*{0.2cm}
    \includegraphics[width=0.24\textwidth]{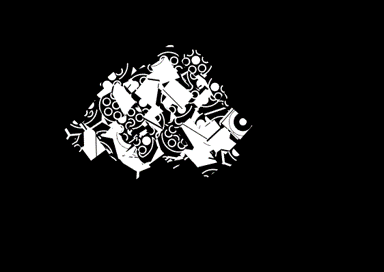}
    \includegraphics[width=0.24\textwidth]{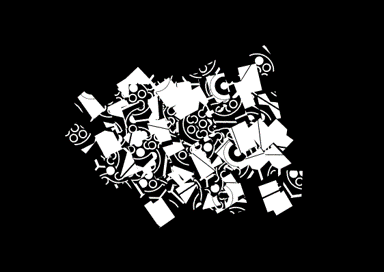}
    \includegraphics[width=0.24\textwidth]{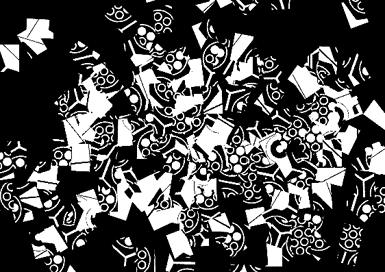}
    \includegraphics[width=0.24\textwidth]{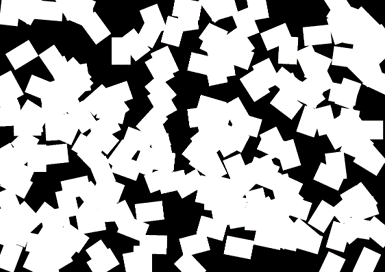}
    % \vspace{-0.1cm}
    \\
    \makebox[0.24\textwidth]{Net a}
    \makebox[0.24\textwidth]{Net b}
    \makebox[0.24\textwidth]{Net c}
    \makebox[0.24\textwidth]{Mask}
    \vspace{0.1cm}
    \\
    \includegraphics[width=0.24\textwidth]{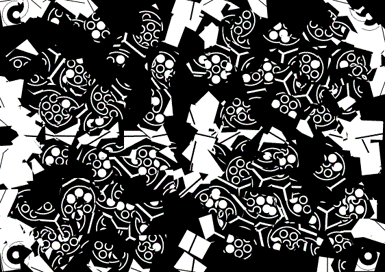}
    \includegraphics[width=0.24\textwidth]{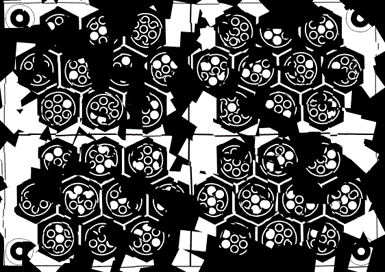}
    \includegraphics[width=0.24\textwidth]{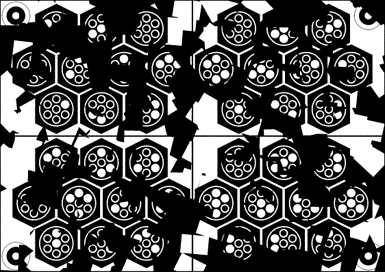}
    \includegraphics[width=0.24\textwidth]{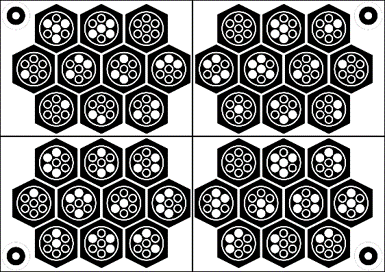}\\
    \makebox[0.24\textwidth]{Noisy Stitched Image}
    \makebox[0.24\textwidth]{Ours}
    \makebox[0.24\textwidth]{Ground truth}
    \makebox[0.24\textwidth]{Reference}
    \vspace{0.3cm}
    \\
    \includegraphics[width=0.24\textwidth]{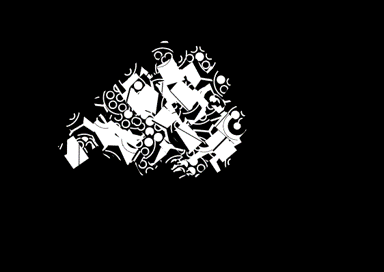}
    \includegraphics[width=0.24\textwidth]{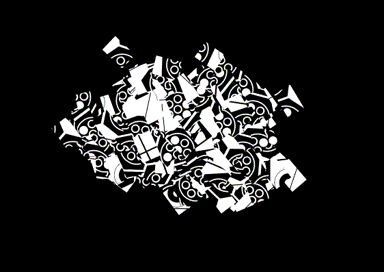}
    \includegraphics[width=0.24\textwidth]{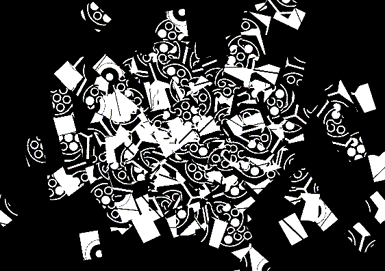}
    \includegraphics[width=0.24\textwidth]{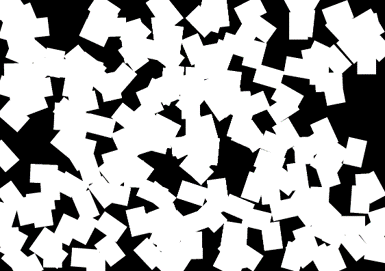}\\
    \makebox[0.24\textwidth]{Net a}
    \makebox[0.24\textwidth]{Net b}
    \makebox[0.24\textwidth]{Net c}
    \makebox[0.24\textwidth]{Mask}
    \vspace{0.1cm}
    \\
    \includegraphics[width=0.24\textwidth]{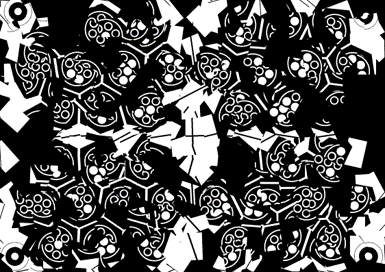}
    \includegraphics[width=0.24\textwidth]{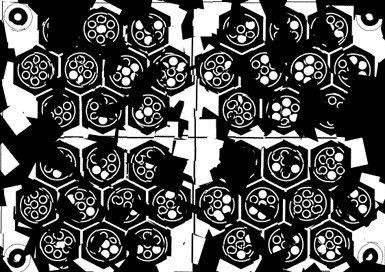}
    \includegraphics[width=0.24\textwidth]{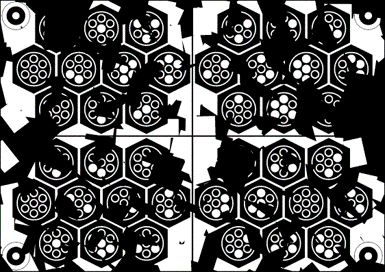}
    \includegraphics[width=0.24\textwidth]{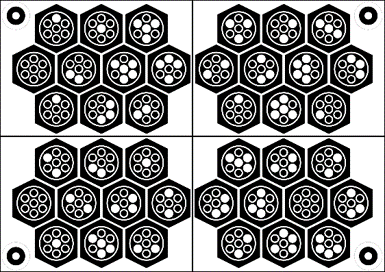}\\
    % \makebox[0.24\textwidth]{Net a}
    % \makebox[0.24\textwidth]{Net b}
    % \makebox[0.24\textwidth]{Net c}
    \makebox[0.24\textwidth]{Noisy Stitched Image}
    \makebox[0.24\textwidth]{Ours}
    \makebox[0.24\textwidth]{Ground truth}
    \makebox[0.24\textwidth]{Reference}
    \\
    \vspace{-0.7em}
    \caption{
    The first two rows correspond to the first example, and the bottom two rows correspond to the second example. 
    The visual comparisons demonstrate that our results exhibit a more coherent and consistent structure.}
    \vspace{-0.4cm}
    \label{fig:comparison}
\end{figure*}

To create the masks for our dataset, we generate a blank canvas with the same dimensions as the synthetic images. The snapshots are then patched onto the canvas, but instead of capturing detailed visual information, we focus on identifying the patched and unpatched regions. The areas corresponding to the snapshots are marked in white (patched), while the rest of the canvas remains black (unpatched). This binary mask explicitly represents where the snapshots are located on the image and captures their spatial orientation and rotation (Mask figure shown in Fig.~\ref{fig:comparison}). This global positional information is crucial for the network to understand the overall layout and context of the snapshots in relation to the entire scene.

\section{Multi-Modals Image Assembly System}

% \vspace*{0.1cm}

The pipeline (Fig.~\ref{fig:overview}) consists of a data processing stage, a deep learning network for coordinate prediction, and an image stitching process to produce the final image.

\subsection{Problem Formulation:}
Our network aims to predict coordinates ${C}_{predicted}$ from noisy coordinates ${C}_{noisy}$ captured by robots, where noise is introduced by environmental factors such as drift and random rotation. ${C}_{predicted}$ is expected to be as close as possible to ${C}_{true}$. To achieve this, we propose a multi-modal framework that integrates visual data and coordinate information to map noisy coordinates to the true ones. Given noisy coordinates ${C}_{noisy} = \{x_{noisy}, y_{noisy}, \theta_{noisy}\}$, corresponding snapshots $S$ (Sec.~\ref{sec:snapshot_generation}), and binary masks $M$ (Sec.~\ref{sec:mask_generation}), the task is to predict the accurate coordinate ${C}_{predicted} = \{x_{true}, y_{ture}, \theta_{true}\}$. This is formulated as a regression problem where the goal is to minimise the error between the predicted and true coordinates. The learned function $f$ (our model) is expressed as:
\begin{equation}
{C}_{predicted} = f(C_{noisy}, S, M).
\end{equation}

\subsection{Multi-Modal Architecture:}
The proposed network architecture is designed to handle multi-modal input, integrating information from snapshots, masks, and noisy coordinates to predict accurate, noise-free coordinates. Snapshots provide detailed local visual information, while masks offer global positional context, both of which are essential for enhancing the representation ability to predict coordinates effectively.
% The proposed network is end-to-end.
We introduce our network in three stages: Feature Extraction, Feature Fusion, and Coordinate Prediction. 

\textbf{Feature Extraction:}
% Specifically, 
Given the input of snapshots $S$, masks $M$ and coordinates $C_{noisy}$, we construct two encoders, ${Enc}_{l}$ and ${Enc}_{g}$ for $S$ and $M$, respectively. ${Enc}_{l}$ applies a depth-wise convolution~\cite{chollet2017xception} followed by a ReLU layer to embed $S$. This design choice is motivated by the nature of the snapshots: after randomly sampling 221 snapshots, each one can have a different pattern due to positional and rotational variations, making them anisotropic with respect to each other. In these cases, conventional convolution operations — where kernels process features across all input channels — are not optimal, as they risk blending distinct patterns between snapshots, failing to capture their unique characteristics. Instead, depth-wise convolutions apply separate filters to each channel independently, preserving the individuality of each snapshot’s features while maintaining computational efficiency. Following the depth-wise embedding, our encoder downsamples the features for the fusion step. 

For ${Enc}_{g}$, which processes the mask $M$, we apply three $\{Conv-ReLU\}$ blocks. The first one performs feature embedding, while the second and third blocks apply downsampling to progressively reduce the spatial resolution while retaining the global information from the visual features. At the end of both encoders, global average pooling is applied to reduce the dimensionality of the visual features for a compact representation, which is appropriate given that the target output $C$ is relatively low-dimensional. After global average pooling, we have the features $f_l$ and $f_g$ extracted from ${Enc}_{l}$ and ${Enc}_{g}$, respectively.

For the noisy coordinates $C_{noisy}$, we stack four fully connected linear layers, each activated by ReLU, to embed the coordinate features into a higher-dimensional space. The embedded feature is denoted as $f_c$. 

\textbf{Feature Fusion and Coordinate Prediction:}
Once we extract $f_l$, $f_g$ and $f_c$, we fuse them to create a unified feature representation $f_{combined}$. This fused representation integrates local visual information, global positional data, and the embedded noisy coordinates, providing a rich feature space that enhances the representation learning capability of the model, leading to more accurate coordinate prediction. Finally, $f_{combined}$ is passed through three fully connected layers, each activated by ReLU, to progressively refine the fused features. The network outputs final coordinates \{x,y,$\theta$\}.

\subsection{Image Assembly}
To assemble the final image, we process each snapshot by applying rotation, positional alignment, and compositing onto a larger canvas. Based on the provided coordinates $\{x,y, \theta\}$, the snapshot is then rotated to correct for any misalignment. Spatial alignment is achieved by positioning the rotated snapshot at the predicted coordinates by compositing the snapshot onto the canvas.
\begin{table*}[htbp]
% \captionsetup[table]{skip=4pt}
% \vspace{-0.2cm}
\vspace*{0.2cm}
\caption{Ablation studies of each component.For MSE, MAE and $R^2$, the values separated by `/' correspond to $x$, $y$, and $\theta$, respectively.}
% \vspace{-0.2cm}
	\centering
        \resizebox{\textwidth}{!}
        {
		\begin{tabular}{@{\extracolsep{3pt}} c c c c c  | c c c c@{}}
				\hline\hline
                    \multicolumn{1}{c}{\multirow{2}{*}{Net}} & 
				  \multicolumn{4}{c}{Components} &\multicolumn{4}{c}{Metrics}\\
					& Coor & Snapshots & Mask & Noisy Stitched Image  & MSE$\downarrow$ & MAE$\downarrow$ & $R^2\uparrow$ & IoU$\uparrow$\T\B\\
				\hline\hline
        \textbf{\textcolor{black}{(a)}}  & \checkmark & & & &  0.0897 / 0.0901 / 0.3639 &  0.2514 / 0.2520 / 0.5092 &  -0.1047 / -0.1215 / -0.0821 & 18.94\%  \\
        \textbf{\textcolor{black}{(b)}} &\checkmark &\checkmark  & & & 0.0807 / 0.0807 / 0.3328 &  0.2395 /  0.2405 / 0.4878 & 0.0060 / -0.005 / 0.0115 & 20.32\% \\
        \textbf{\textcolor{black}{(c)}} &\checkmark &\checkmark & &\checkmark &  0.1384 / 0.1243 / 0.5250 &   0.3007 / 0.2874 / 0.5870 &  -0.7023 / -0.5527 / -0.5624  & 43.47\% \\
        \textbf{\textcolor{black}{Ours}} &\checkmark &\checkmark  &\checkmark & &  \textbf{0.0019} /  \textbf{0.0022} / \textbf{0.0091} &  \textbf{0.0156} /  \textbf{0.0163} / \textbf{0.0329} & \textbf{0.9503} / \textbf{0.9411} / \textbf{0.9465} & \textbf{87.97\%}  \B\\
        \hline
        \hline 
        \end{tabular}
        }
\label{tab:ablation}
\vspace{-0.5cm}
\end{table*}

\section{Experiments}

All experiments are conducted on SFP10 (Sec.~\ref{sec:data}).
% Our model is trained on the proposed SFP10 dataset, with 10,000 sets of synthetic data (Sec.~\ref{sec:data}). Similarly, 1,000 synthetic images are generated for evaluation.

\subsection{Implementation Details}
All data processing tasks - i.e. generating synthetic images, snapshots, masks, and noisy coordinates, were conducted on an Intel Core i9 10 Core Processor i9-10900X (3.7GHz) 19.25MB Cache. For model training and testing, all experiments were carried out on a single Nvidia 3070Ti GPU. During training, we calculate the loss for both the location $(x,y)$ and orientation $\theta$ using Mean Squared Error (MSE). The total loss is defined as:
\begin{equation}
    L_{total}(C_{predicted}, C_{true}) = L_{x,y} + 0.05 L_{\theta}.
\end{equation}
We used the Adam optimizer with $\beta_1 = 0.9$ and $\beta_2 = 0.999$. The learning rate was set to $1\times10^{-4}$, with a batch size of 24 to balance memory efficiency and model convergence. 

\subsection{Evaluation Metrics}

To evaluate the accuracy of the predicted coordinates, we use Mean Squared Error (MSE), Mean Absolute Error (MAE), and the coefficient of determination ($R^2$). MSE measures the average squared difference between predicted and true coordinates, penalising larger errors more heavily, which is ideal for detecting significant deviations. MAE, on the other hand, calculates the average absolute error, and offers a straightforward interpretation of the overall accuracy. 
$R^2$ assesses how well the model's predictions fit the true coordinates. We also considered the issue of angle rotations differing by integer multiples of 360 degrees, which could lead to identical rotations but distort error measurement. However, in our experiment, the rotation of all snapshots is limited to less than 2.5 radians, which simulates real-world scenarios from our observation. Therefore, the above metrics are effective in evaluating our method.

To evaluate our assembled images, we use Intersection over Union (IoU) \cite{everingham2010pascal} to assess the overlap between the mask with the ground truth coordinates and the mask with the predicted coordinates. While IoU does evaluate denoising performance, it is not a precise indicator. Even with noisy coordinates, the IoU can be high (above 80\%) because the noisy snapshots still overlap significantly with the ground truth region. As the model reduces noise, the IoU increases, ideally reaching 100\% when all noise is removed. Therefore, while IoU helps track improvement, it must be interpreted alongside other metrics for a more accurate evaluation.

\subsection{Evaluation and Discussion}
To evaluate each component in our proposed method, we conduct four configurations: Net (a), (b), (c) and our method. The key difference between these configurations lies in their input components. Net (a) uses only $[\textit{coordinates}]$, while Net (b) incorporates $[\textit{coordinates, snapshots}]$. Net (c) inputs $[\textit{coordinates, snapshots, noisy assembled image}]$. Our proposed method employs $[\textit{coordinates, snapshots, and masks}]$.
% , where the mask provides critical global positional information, enabling more accurate predictions of both position and orientation.

As shown in Table~\ref{tab:ablation}, Net (a) and Net (b) demonstrate that simply using noisy coordinates or noisy coordinates with snapshots fails to provide sufficient information to accurately predict the correct coordinates, resulting in poor assembly of snapshots.
In contrast, our method integrates input noisy coordinates, snapshots, for local visual information, and masks, for global context, to accurately predict $\{x,y,\theta\}$, resulting in well-aligned outputs. As shown in Table~\ref{tab:ablation}, incorporating the mask provides the global positional information necessary for the model to outperform other configurations, leading to better alignment and overall higher accuracy.

Net (c) explores the feasibility of using the noisy assembled image (with noisy coordinates) as an alternative to the mask for providing global information (replacing the mask as input for $Enc_g$). Even though it tends to predict the sparse $\{x, y\}$ location, it is unable to learn the global context, resulting in misalignment in the final assembled image (shown in the Fig.~\ref{fig:comparison}). This limitation could stems from redundant and inaccurate visual information provided by the noisy assembled image, which makes it challenging for the model to predict the $\theta$ accurately.

% While this approach is also able to predict $(x,y)$, the redundant and noisy information in the composite image makes it challenging for the model to predict $\theta$ accurately. This results in misalignment of the final image, particularly degrading the prediction of $\theta$ compared to our method, as reflected in the metrics.

% \textcolor{red}{\textbf{$\uparrow$ Unfinished, Chris will finish it.}}

% \section{Results \& Discussions}

\section{CONCLUSIONS}
In this work, we have proposed a synthetic dataset SFP10, and a novel approach for improving visual monitoring in noisy underwater environments using a swarm of micro-robots. Our proposed pipeline integrates data simulation, a multi-modal deep learning network for coordinate prediction, and image reassembly to address challenges posed by environmental disturbances, such as drift and rotation. By incorporating local visual information from snapshots and global positional context from masks, our method significantly enhances the precision of both coordinate prediction and image alignment.

Although our approach has shown promising results with synthetic data, there remains a domain gap between real-world and simulated data. Future work will focus on closing this gap by exploring techniques to minimise discrepancies between synthetic data and real-world scenes. This includes exploring advanced techniques such as domain adaptation \cite{atapour2019generative}, transfer learning \cite{liu2020skill} and data augmentation \cite{jackson2019style} to better generalise from synthetic to real-world data. Additionally, incorporating real-world datasets into the training process \cite{atapour2019complete} and leveraging unsupervised \cite{liu2022deep} or semi-supervised \cite{cheng2014semi} learning approaches may further bridge the domain gap. This will ensure the robustness and applicability of our method in real-world underwater environments. Additionally, we plan to extend the system's scalability and optimise it for real-time performance \cite{shen2019real} to broaden its use in other extreme environments such as pipeline inspections.

\bibliography{Main}
\bibliographystyle{ieeetr}

\end{document}